\documentclass{article}
\usepackage[final]{neurips_2019}
\usepackage{graphicx,calc,color,subfigmat,amssymb,amsmath,mathrsfs,dsfont,epstopdf}

\usepackage[utf8]{inputenc} 
\usepackage[T1]{fontenc}    
\usepackage{hyperref}       
\usepackage{url}            
\usepackage{booktabs}       
\usepackage{amsfonts}       
\usepackage{nicefrac}       
\usepackage{microtype}      
\usepackage{natbib}
\newcommand{\Pro}{\mathrm{Pr}}
\newtheorem{theorem}{Theorem}

\newcommand{\pardef}{\stackrel{\rm def}{=}}

\title{Signal to Noise Ratio Loss Function}

\author{%
  Ali Ghobadzadeh\\
  Huawei Technologies\\
  Toronto, ON \\
  \texttt{ali.ghobadzadeh1@huawei.com} \\
  \And
  Amir Lashkari\\
  Huawei Technologies\\
  Toronto, ON \\
  \texttt{amir.lashkari1@huawei.com}
}

\begin{document}

\maketitle

\begin{abstract}
This work proposes a new loss function targeting classification problems, utilizing a source of information overlooked by cross entropy loss. First, we derive a series of the tightest upper and lower bounds for the probability of a random variable in a given interval. Second, a lower bound is proposed for the probability of a true positive for a parametric classification problem, where the form of probability density function (pdf) of data is given. A closed form for finding the optimal function of unknowns is derived to maximize the probability of true positives. Finally, for the case that the pdf of data is unknown, we apply the proposed boundaries to find the lower bound of the probability of true positives and upper bound of the probability of false positives and optimize them using a loss function which is given by combining the boundaries. We demonstrate that the resultant loss function is a function of the signal to noise ratio both within and across logits. We empirically evaluate our proposals to show their benefit for classification problems.
\end{abstract}

\section{Introduction}\label{sec:Introduction}
The cross entropy loss function has had broad success in the problem domain of classification 
\cite{mannor2005cross,girshick2014rich,moore2010intelligent,kline2005revisiting, lecun1989backpropagation, vaillant1994original}. 
In some important applications, cross entropy alone is not sufficient to achieve state-of-the-art results. 
For example, in the field of face recognition, the best results are achieved by combining cross entropy loss with discriminative losses, such as center loss
\cite{Wen16}, triplet loss \cite{Schroff15}, angular softmax loss \cite{Liu17}, and large margin softmax loss \cite{Liu16}. The aforementioned losses improve the distance of features and logits through augmenting the cross entropy loss. In this work, we propose a loss function to improve upon the discriminative power of cross entropy loss. 

There exists a tight relationship between the cross entropy and the maximum likelihood estimation loss functions, as both losses maximize the probability of logits for each class. 
These losses have been used across a variety of models with strong results 
\cite{krizhevsky2009learning, simonyan2014very, he2016deep}. 
However, the formulation of these losses provides feedback based solely on the positive class of a given sample. They do not incorporate the intuition that, when a sample is of a particular class, it is not any of the other classes, and its predicted likelihood of belonging to those other classes should be low. Guided by this intuition, this work seeks to establish a framework to explicitly incorporate this new source of supervision, which provides more information during training and a corresponding improvement in accuracy. 
%

Several loss functions have previously been proposed to extend or improve on cross entropy. 
Focal Loss is an augmented cross entropy loss function that is applied to imbalanced datasets \cite{lin2017focal}. In \cite{cho2015hessian}, a convex approximation of a variation form of cross entropy is applied to recurrent networks. 
In face recognition applications, several approaches have been tried to improve discriminatory power 
\cite{Taigman14,Sun14, Wen16,Liu16,Liu17}. 
These include modified loss functions, 
such as the center loss \cite{Wen16}, triplet loss \cite{Schroff15}, angular softmax loss \cite{Liu17}, and large margin softmax loss \cite{Liu16}. 
These loss functions are directly applied to feature representations to maximize inter-class distances and minimize intra-class distances between them. 
The triplet loss seeks to improve the learned Euclidean embedding space based on the relative inter- and intra-class distances \cite{Schroff15}. 
Additionally, cosine similarity works well as a metric for the clustering task, hence, the angular margin \cite{Liu17} and large margin cosine loss \cite{Wang18} have been pursued as objective functions. 
Compared to the Euclidean margin \cite{Wen16,Chopra05,Hoffer15}, the angular margin performs better when the model is trained with softmax. While these losses seek to gain enhanced discriminatory power, they overlook the information presented by negative annotations which our proposed loss utilizes.

To develop the proposed approach, it is necessary to first identify the tightest possible upper and lower bounds for the probability of true positives and false positives for each logit in terms of its mean and variance. Then, we maximize the lower bound of probability for true positives and minimize the upper bound of probability for false positives. 


In the case that the family of probability of density functions (pdf) under each class is given, the classifier is found by replacing the estimation of unknown parameters of pdf and placing them into the likelihood function \cite{lehmann2006testing}. Although the likelihood function is the optimal function for classification with completely known pdfs, introducing the estimation of unknown parameters into it may negatively affect it \cite{lehmann2006testing}. Furthermore, estimating unknown parameters and inserting them into the likelihood function, necessarily, is not the optimal method when the pdf has unknown parameters. Hence, we propose a method to find the optimal function by maximizing a lower bound of the probability of true positives. 

We put forward two methods for applying the proposed method during model training, and provide a quantitative comparison to cross entropy on its own. We report on the MNIST dataset using a LeNet architecture \cite{lecun1998gradient}, as well as CIFAR-10 and CIFAR-100 using ResNet-22 \cite{he2016deep}. In all cases evaluated, the addition of the proposed loss improves performance significantly.

The contributions of this paper are: First, we derive the tightest upper and lower bounds for the probability of a random variable in a given interval. 
Second, based on the proposed boundaries, we provide a lower bound for the probability of a true positive for a parametric classification problem. 
A parametric classification problem is a problem where the family of the pdf is known, while the pdf has some unknown parameters. 
We prove a closed form for finding the optimal function of unknowns to maximize the probability of true positives. 
Finally, we apply the proposed boundaries to find the lower bounds of probability of true positives and upper bound of probability of false positives and optimize them using a loss function which is given by combining the boundaries. 
Interestingly, the loss function is a function of the ratio of mean and variance, which we identify as the signal to noise ratio of within and between logits. 
To verify the proposed approach, it is applied to non-parametric problems, where the pdf of data is unknown. 
To evaluate the proposed method, LeNet and ResNet-22 are trained using the proposed loss along with cross entropy in two different strategies. 



\section{Upper and Lower bounds for characteristics function of random variable}\label{sec_2}
In this section, we derive the tightest upper and lower bounds for a random variable with mean $\mu$ and variance $\sigma^2$. 
Given $x$, a random variable with mean $\mu$ and variance $\sigma^2$, we show that for any given threshold $\eta$, $\Pro(x>\eta)$ and $\Pro(x<\eta)$ can be bounded using a closed form of $\mu$ and $\sigma^2$. 
\begin{theorem}\label{Theo_1}
Let $x$ be a random variable with mean $\mu$ and variance $\sigma^2$, then for any arbitrary $\eta$, the tightest upper bound of $\Pro(x>\eta)$ in terms of $\mu$ and $\sigma^2$ is $\frac{\sigma^2}{\sigma^2 + (\eta - \mu)^2}$, for $\eta\geq\mu$ and $1$ for $\eta<\mu$.
\end{theorem}
\textbf{Proof:} First, we consider $\eta<\mu$. Assume that the random variable $x$ has the following probability density function (pdf):  $f(x)=\frac{\epsilon}{2}\delta(x-\mu+\frac{\sigma}{\sqrt{\epsilon}})+(1-\epsilon)\delta(x-\mu)+\frac{\epsilon}{2}\delta(x-\mu-\frac{\sigma}{\sqrt{\epsilon}})$, where $\epsilon\in(0,1)$ and $\delta(\cdot)$ is the delta function.\footnote{$\delta(x) = 1$ iff $x =0$, and $\delta(x) = 0$ at all other points.} The mean and variance of this pdf are $\mu$ and $\sigma^2$ respectively, while $\Pro(x>\eta)=1-\frac{\epsilon}{2}$. As $\epsilon$ can take any arbitrary small positive value, the tightest possible bound is $1$. Therefore, there exists a pdf that satisfies the upper bound.

To prove the case where $\eta \geq \mu$, without loss of generality, we first find the bound for an $x$ with $\mu = 0$ and $\sigma^2 = 1$. Then, we extend the result to the general case. 

Let $\tilde{x}$ denote a random variable with a mean of zero and variance equal to one. We therefore have $\tilde{\eta}=\frac{\eta-\mu}{\sigma}>0$. The tightest upper bound for $\Pro(\tilde{x}>\tilde{\eta})$ is given by solving the following maximization problem
\begin{eqnarray}\label{OPT_EQ_1}
\sup_{f(\tilde{x})}\limits & \int_{\eta}^{\infty}f(\tilde{x})\mathrm{d}\tilde{x}\\\label{OPT_EQ_1b}
\mathrm{s.t.} & \int_{-\infty}^{\infty}\tilde{x}f(\tilde{x})\mathrm{d}\tilde{x} = 0,\\\label{OPT_EQ_1c} 
&\int_{-\infty}^{\infty}\tilde{x}^2f(\tilde{x})\mathrm{d}\tilde{x} = 1.
\end{eqnarray}
The optimization problem is on the set of all possible pdfs that satisfy (\ref{OPT_EQ_1b}) and (\ref{OPT_EQ_1c}). Defining $a\pardef \int_{\eta}^{\infty}f(\tilde{x})\mathrm{d}\tilde{x}>0$, $b \pardef \int_{-\infty}^{\eta}f(\tilde{x})\mathrm{d}\tilde{x}>0$, $\tilde{x}_a\pardef \Big(\frac{1}{2}\Big((\frac{1}{b}-\frac{1}{a}+1)^2+\frac{4}{a}\Big)^{1/2}-(\frac{1}{b}-\frac{1}{a}+1)\Big)^{1/2}\geq 0$ and $\tilde{x}_b \pardef \frac{\tilde{x}_a^2-\frac{1}{a}}{\tilde{x}_a}\leq 0$
we can rewrite (\ref{OPT_EQ_1}), (\ref{OPT_EQ_1b}) and (\ref{OPT_EQ_1c}) as follows: 
$\sup a$ subject to $a\tilde{x}_a + b\tilde{x}_b = 0$, $a\tilde{x}_a^2 + b\tilde{x}_b^2 = 1$, $a + b=1$ and $a\geq 0, b \geq 0$, which is a linear optimization problem. We will now demonstrate that by adding two more constraints to the optimization problem, $\tilde{x}_b \leq 0$ and $\tilde{x}_a\geq \tilde{\eta}$, we arrive at a unique solution.

We represent the pdf $f(\tilde{x})$ 
as a new pdf composed of two impulses,  $\overline{f}_{\tilde{x}}(\tilde{x})=a\delta(\tilde{x}-\tilde{x}_a) + b\delta(\tilde{x}-\tilde{x}_b)$, where $\Pro(\tilde{x}\geq\eta) = a$. 
There are two possibilities, either $\tilde{x}_a \geq \eta$ or $\tilde{x}_a < \eta$. 
For any pdf $\overline{f}_{\tilde{x}}(\tilde{x})$ with $\tilde{x}_a < \eta$, there exists another pdf in form $(a+\Delta)\delta(\tilde{x}-\tilde{\eta}) + (b-\Delta)\delta(\tilde{x}-\tilde{x}'_b)$, where $\Delta\in[0,1]$. 
In other words, for $f_{\tilde{x}}(\tilde{x})$ and its representative pdf $\overline{f_{\tilde{x}}}(\tilde{x})$ with $\tilde{x}_a < \eta$, there exists another pdf with two delta functions such that $\Pro(\tilde{x}\geq\eta)$ is greater than $\int_{\tilde{\eta}}^{\infty}f_{\tilde{x}}(\tilde{x})\mathrm{d}\tilde{x}$. 
The validity of this claim can be shown by defining $\Delta\pardef\frac{b(1-a\tilde{\eta}^2)}{\tilde{\eta}^2+1}$. To see this, consider the following scenarios \\
1) $\Delta>0$; since $a\tilde{\eta}^2= \tilde{\eta}^2\int_{\tilde{\eta}}^{\infty}f_{\tilde{x}}(\tilde{x})\mathrm{d}\tilde{x}\leq \int_{\tilde{\eta}}^{\infty}\tilde{x}^2 f_{\tilde{x}}(\tilde{x})\mathrm{d}\tilde{x}<1$, the last inequality is given by the definition of variance for $f_{\tilde{x}}(\tilde{x})$. Thus $a\tilde{\eta}^2<1$ or equivalently $\Delta>0$. \\
2) $a+\Delta\in[0,1]$; since $a+\Delta= \frac{a^2\tilde{\eta}^2+1}{\tilde{\eta}^2+1}$.\\
3) $b-\Delta\in[0,1]$; since $a+\Delta\in[0,1]$.\\
4) $\tilde{x}_b'<0$ since the mean of $(a+\Delta)\delta(\tilde{x}-\tilde{\eta}) + (b-\Delta)\delta(\tilde{x}-\tilde{x}'_b)$ is zero.\\

Based on this claim, to find the supremum of $\int_{\eta}f_{\tilde{x}}(\tilde{x})\mathrm{d}\tilde{x}$, we need to find the supremum over all pdfs with two delta functions, such that the location of one delta function is negative and the location of the other is greater than or equal to $\tilde{\eta}$.
Thus, the optimization problem in Equation \ref{OPT_EQ_1} becomes
\begin{align}\label{OPT_EQ_2}
\sup \ \ \ &  a\\
\nonumber \mathrm{s.t.}\ \ \  & a\tilde{x}_a + b\tilde{x}_b = 0, a\tilde{x}_a^2 + b\tilde{x}_b^2 = 1,\\
\nonumber & a + b=1, a\geq 0, b \geq 0, \tilde{x}_b \leq 0,\tilde{x}_a\geq \tilde{\eta},
\end{align}
where the last inequality is given by $\Pro(\tilde{x}\geq \tilde{\eta})>0$, so $\tilde{x}_a\geq \tilde{\eta}$. From $a\tilde{x}_a + b\tilde{x}_b = 0$, $a\tilde{x}_a^2 + b\tilde{x}_b^2 = 1$, and $a + b=1$, we have $a=\frac{1}{1 + \tilde{x}_a}$. Since $\tilde{x}_a\geq \tilde{\eta}$, we see $a\geq \frac{1}{1 + \tilde{\eta}^2}$, and the tightest bound for $\Pro(\tilde{x}\geq \tilde{\eta})$ is $\frac{1}{1 + \tilde{\eta}^2}$. For any arbitrary random variable $x$ with mean $\mu$ and variance $\sigma^2$, there exists $\tilde{x}$ with mean zero and variance one such that $\tilde{x}=\frac{x - \mu}{\sigma}$. From this, one can see that $\Pro(x\geq \eta)=\Pro(\tilde{x}\geq \tilde{\eta})\leq \frac{1}{1 + (\frac{\eta-\mu}{\sigma})^2}=\frac{\sigma^2}{\sigma^2 + (\eta-\mu)^2}$.
$\blacksquare$\\

To find the tightest upper bound for $\Pro(x\leq \eta)$, from Theorem \ref{Theo_1}, we have
\begin{align}\label{CDF_upper_0}
\Pro(x\leq \eta)=\Pro(-x\geq -\eta)\leq \left\{\begin{array}{lllll}
1, & \eta > \mu,\\
\frac{\sigma^2}{\sigma^2 + (\eta-\mu)^2}, & \eta \leq \mu.
\end{array}
\right.
\end{align}
The last inequality is given by applying Theorem \ref{Theo_1} on a random variable with mean $-\mu$ and variance $\sigma^2$ and threshold $-\eta$. Moreover, using Theorem \ref{Theo_1}, the tightest lower bound for $\Pro(x\leq \eta)$ is given as follow
\begin{eqnarray}\label{CDF_lower_0}
\Pro(x\leq \eta) = 1 - \Pro(x\geq \eta)\geq \left\{\begin{array}{lllll}
\frac{(\eta-\mu)^2}{\sigma^2 + (\eta-\mu)^2},& \eta \geq \mu,\\
0,& \eta < \mu.
\end{array}
\right.
\end{eqnarray}
The feasible interval for a cumulative distribution function (CDF) given by (\ref{CDF_upper_0}) and (\ref{CDF_lower_0}) is
\begin{align}\label{CDF_feasible_0}
\left\{\begin{array}{lllll}
\frac{(\eta-\mu)^2}{\sigma^2 + (\eta-\mu)^2}, & \eta \geq \mu,\\
0, & \eta < \mu.
\end{array}
\right.
\leq F_{x}(\eta)
\leq
\left\{\begin{array}{lllll}
1, &\eta > \mu,\\
\frac{\sigma^2}{\sigma^2 + (\eta-\mu)^2}, & \eta \leq \mu,
\end{array}
\right.
\end{align}
where $F_{x}(\eta)$ is the CDF of $x$ at $\eta$. 


\begin{theorem}\label{Theo_2}
For any $x$ with mean $\mu$ and variance $\sigma^2$, the tightest upper bound for $\Pro(x \leq \eta_1, x \geq \eta_2)$ is given by
\begin{align}\label{Theo_1_EQ_0}
\Pro(x \leq \eta_1; x \geq \eta_2)\leq 
\left\{\begin{array}{lllll}
\max\{\frac{\sigma^2}{|\eta_1-\mu|(\eta_2-\mu)}, \frac{\sigma^2}{\sigma^2 + (\min\{|\eta_1-\mu|,(\eta_2-\mu)\})^2}\}, & |\eta_1-\mu|(\eta_2-\mu)\geq \sigma^2,\\
1, & |\eta_1-\mu|(\eta_2-\mu)< \sigma^2,
\end{array}
\right.
\end{align}
where $\eta_1<\mu$ and $\eta_2>\mu$. Moreover, if $\eta_1>\mu$ and $\eta_2>\mu$, or $\eta_1<\mu$ and $\eta_2<\mu$, then the tightest bound is $1$.
\end{theorem}
\textbf{Proof: }
Defining $\tilde{x}=\frac{x-\mu}{\sigma}$, $\tilde{\eta}_1=\frac{\eta_1-\mu}{\sigma}$ and $\tilde{\eta}_2=\frac{\eta_2-\mu}{\sigma}$, we have $\Pro(x \leq \eta_1, x \geq \eta_2) = \Pro(\tilde{x} \leq \tilde{\eta}_1, \tilde{x} \geq \tilde{\eta}_2)$. To prove (\ref{Theo_1_EQ_0}), first assume that $|\eta_1-\mu|(\eta_2-\mu)< 1$, or equivalently, $|\tilde{\eta}_1|\tilde{\eta}_2<1$, where $\tilde{\eta}_2>0$ and $\tilde{\eta}_1<0$. By defining $\lambda\pardef k\max\{|\tilde{\eta}_1|,\tilde{\eta}_2,1\}$, where $k=-1$ if $|\tilde{\eta}_1|>\tilde{\eta}_2$ and $|\tilde{\eta}_1|>1$ else $k=1$, and pdf $f_{\tilde{x}}(\tilde{x})=\frac{\lambda^2}{\lambda^2+1}\delta(\tilde{x}+\frac{1}{\lambda})+\frac{1}{\lambda^2+1}\delta(\tilde{x}-\lambda)$, we have that the mean and variance of $\tilde{x}$ are zero and one, respectively, and $\Pro(x \leq \tilde{\eta}_1, \tilde{x} \geq \tilde{\eta}_2)=1$.

To prove the rest of theorem, we apply the same procedure as the proof of Theorem \ref{Theo_1}. 
For any pdf $f_{\widetilde{x}}(\widetilde{x})$ with zero mean, variance one, and $P_{\tilde{\eta}_1,\tilde{\eta}_2}\pardef\Pro(\tilde{x}\leq\tilde{\eta}_1; \tilde{x}\geq\tilde{\eta}_1)$, there exists a pdf $\overline{f}_{\widetilde{x}}(\widetilde{x})=c\delta(\tilde{x}-\tilde{x}_c)+b\delta(\tilde{x})+a\delta(\tilde{x}-\tilde{x}_a)$ with mean zero and variance one such that $P_{\tilde{\eta}_1,\tilde{\eta}_2} = a+c$. In a similar manner to the proof of Theorem \ref{Theo_1}, by defining $c\pardef\int_{-\infty}^{\tilde{\eta}_1}f_{\tilde{x}}(\tilde{x})\mathrm{d}\tilde{x}$, $a\pardef\int_{\tilde{\eta}_2}^{\infty}f_{\tilde{x}}(\tilde{x})\mathrm{d}\tilde{x}$, $b\pardef\int_{\tilde{\eta}_1}^{\tilde{\eta}_2}f_{\tilde{x}}(\tilde{x})\mathrm{d}\tilde{x}$, $\tilde{x}_a\pardef \Big(\frac{1}{2}\big(\sqrt{(\frac{1}{c}-\frac{1}{a}+1)^2+\frac{4}{a}}-(\frac{1}{c}-\frac{1}{a}+1)\big)\Big)^{1/2}\geq 0$, and $\tilde{x}_c = \frac{1}{\tilde{x}_a}(\tilde{x}_a^2-\frac{1}{a})\leq 0$, the claim is proven.\footnote{This substitution preserves the mean and variance, without any change.} 
In the relationship between $\tilde{x}_c$ and $\tilde{\eta}_1$, as well as $\tilde{x}_a$ and $\tilde{\eta}_2$, we have four scenarios: 
(1) $\tilde{x}_c \leq \tilde{\eta}_1$ and $\tilde{x}_a \geq \tilde{\eta}_2$.
(2) $\tilde{x}_c > \tilde{\eta}_1$ and $\tilde{x}_a < \tilde{\eta}_2$, 
(3) $\tilde{x}_c > \tilde{\eta}_1$ and $\tilde{x}_a \geq \tilde{\eta}_2$, 
(4) $\tilde{x}_c \leq \tilde{\eta}_1$ and $\tilde{x}_a < \tilde{\eta}_2$. 

To find the tightest bound, we need to maximize $a+c$ in each case, then select the maximum of all solutions. For the first case, the maximization problem is given by
\begin{align}\label{PROOF_TH_1_EQ_1}
\sup \ \ \ &  a+c\\
\nonumber \mathrm{s.t.}\ \ \  & a\tilde{x}_a + c\tilde{x}_c = 0, a\tilde{x}_a^2 + c\tilde{x}_c^2 = 1,\\
\nonumber & a + b + c=1, a\geq 0, b \geq 0,c \geq 0, \tilde{x}_c \leq \tilde{\eta}_1,\tilde{x}_a\geq \tilde{\eta}_2,
\end{align}
From $a\tilde{x}_a + c\tilde{x}_c = 0, a\tilde{x}_a^2 + c\tilde{x}_c^2 = 1$, $a + b + c=1$, we have $a+c=\frac{1}{|\tilde{x}_c|\tilde{x}_a}$. Since $|\tilde{x}_c| \geq |\tilde{\eta}_1|$ and $\tilde{x}_a \geq \tilde{\eta}_2>0$, the supremum value of $a+c$ is $\frac{1}{|\tilde{\eta}_1|\tilde{\eta}_2}$.

For the second case, without loss of generality, assume that $\tilde{\eta}_2\leq|\tilde{\eta}_1|$. We can replace the the corresponding pdf with $(b-\Delta)\delta(\tilde{x}-\tilde{x}'_{b})+(a+c+\Delta)\delta(\tilde{x}-\tilde{\eta}_2)$, where $\tilde{x}'_{b}>\tilde{\eta}_1$, $(a+c+\Delta)\in[0,1]$ and $(b-\Delta)\in[0,1]$. 
In other words, such pdfs can be replaced by other pdfs such that $P_{\tilde{\eta}_1,\tilde{\eta}_2} = a+b \leq a+c+\Delta$. 
Defining $\Delta=\frac{b(1-(a+c)\tilde{\eta}_2^2)}{1+\tilde{\eta}_2^2}$ and $\tilde{x}'_{b}=\frac{-\tilde{\eta}_2(a+c+\Delta)}{b-\Delta}$, the claim is proven. Therefore, we maximize $a+c+\Delta$ between such pdfs to find an upper bound for $a+b$. 
From Theorem \ref{Theo_1}, the maximum value of $a+b+\Delta$ is $\frac{1}{1+\tilde{\eta}_2^2}$. 
Applying a similar approach of Case 2, for Case 3 and 4, the supremum value of Case 3 and 4 is $\frac{1}{1+\tilde{\eta}_2^2}$. Since the pdf $\frac{(\tilde{\eta}_2 - \epsilon)^2}{(\tilde{\eta}_2 - \epsilon)^2+1}\delta(\tilde{x}+\frac{1}{(\tilde{\eta}_2 - \epsilon)})+\frac{1}{(\tilde{\eta}_2 - \epsilon)^2+1}\delta(\tilde{x}-(\tilde{\eta}_2 - \epsilon))$ belongs to Case 3 for a sufficiently small value of $\epsilon$, the proposed bound is the tightest bound for Case 3.

Following on from this, the supremum of $a+b$ is $\max\{\frac{1}{|\tilde{\eta}_1|\tilde{\eta}_2}, \frac{1}{1+\tilde{\eta}_2}\}$, or equivalently, $\sup (a+b) = \frac{1}{|\tilde{\eta}_1|\tilde{\eta}_2}$ if $|\tilde{\eta}_1|\leq\frac{1 + \tilde{\eta}_2^2}{\tilde{\eta}_2}$ and $\sup (a+b) = \frac{1}{1 + \tilde{\eta}^2_2}$ if $|\tilde{\eta}_1|> \frac{1 + \tilde{\eta}_2^2}{\tilde{\eta}_2}$. By substituting $\tilde{x} = \frac{x-\mu}{\sigma}$, $\tilde{\eta}_1 = \frac{\eta_1-\mu}{\sigma}$, and $\tilde{\eta}_2 = \frac{\eta_2-\mu}{\sigma}$, the proof of (\ref{Theo_1_EQ_0}) is complete.

To prove the remainder of theorem, when $\eta_1>\mu$ and $\eta_2>\mu$, or $\eta_1<\mu$ and $\eta_2<\mu$, consider a random variable $x$ with a pdf $f_{x}(x)=\frac{\epsilon}{2}\delta(x-\mu+\frac{\sigma}{\sqrt{\epsilon}})+(1-\epsilon)\delta(x-\mu)+\frac{\epsilon}{2}\delta(x-\mu-\frac{\sigma}{\sqrt{\epsilon}})$, mean $\mu$, and variance $\sigma^2$. For a sufficiently small value of $\epsilon$, $\Pro(x\leq \eta_1,x\geq \eta_1)=1$.
$\blacksquare$\\

The following theorem proposes the tightest upper bound for $\Pro(\eta_1\leq x\leq \eta_2)$.
\begin{theorem}\label{Theo_3}
The tightest upper bound of $\Pro(\eta_1\leq x\leq \eta_2)$ in terms of mean $\mu$ and variance $\sigma^2$ is $\frac{\sigma^2}{\sigma^2 + (\min{|\eta_1-\mu|,|\eta_2-\mu|})^2}$, if $\eta_1>\mu$ and $\eta_2>\mu$, or $\eta_1<\mu$ and $\eta_2<\mu$. Moreover if $\eta_1<\mu<\eta_2$, the tightest upper bound is one.
\end{theorem}
\textbf{Proof: }
To prove the first part of theorem, without loss of generality, assume that $\eta_1>\mu$ and $\eta_2>\mu$, then $\Pro(\eta_1\leq x\leq \eta_2)\leq  \Pro(\eta_1\leq x)\leq \frac{\sigma^2}{\sigma^2+(\eta_1-\mu)^2}$, where the final inequality comes from Theorem \ref{Theo_1}. Since, for a pdf $\frac{\sigma^2}{\sigma^2 + (\eta_1-\mu)^2}\delta(\frac{x-\eta_1}{\sigma}) + \frac{(\eta_1-\mu)^2}{\sigma^2 + (\eta_1-\mu)^2}\delta(\frac{x-\mu}{\sigma} - \frac{\sigma}{\eta_1-\mu})$, $\Pro(\eta_1\leq x \leq \eta_2)=\frac{\sigma^2}{\sigma^2+(\eta_1-\mu)^2}$, the achieved bound is the tightest one. For $\eta_1<\mu$ and $\eta_2<\mu$, the proof is similar.

To prove the remainder of theorem, consider $f_{x}(x)=\frac{\epsilon}{2}\delta(x-\mu+\frac{\sigma}{\sqrt{\epsilon}})+(1-\epsilon)\delta(x-\mu)+\frac{\epsilon}{2}\delta(x-\mu-\frac{\sigma}{\sqrt{\epsilon}})$, which results in a supremum of $\Pro(\eta_1\leq x \leq \eta_2)$ equal to $1$ if $\eta_1<\mu<\eta_2$. $\blacksquare$\\

From Theorem \ref{Theo_2}, the tightest lower bound for $\Pro(\eta_1\leq x \leq \eta_2)$ is given by
\begin{align}\label{Theo_1_EQ_15}
\Pro(\eta_1 \leq x \leq \eta_2)\geq  \left\{\begin{array}{lllll}
1 - \max\{\frac{\sigma^2}{|\eta_1-\mu|(\eta_2-\mu)},  \frac{\sigma^2}{\sigma^2 + (\min\{|\eta_1-\mu|,(\eta_2-\mu)\})^2}\}, 
|\eta_1-\mu|(\eta_2-\mu)\geq \sigma^2,\\
0 , \ \ \ \ \ \ \ \ \ \ \ \ \ \ \ \ \ \ \ \ \ \ \ \ \ \ \ \ \ \ \ \ \ \ \ \ \ \ \ \ \ \ \ \ \ \ \ \ \ \ \ \ \ \ \ \ \ \ \ \ |\eta_1-\mu|(\eta_2-\mu)< \sigma^2,
\end{array}
\right.
\end{align}
if $\eta_1<\mu<\eta_2$ and if $\eta_1>\mu$ and $\eta_2>\mu$, or $\eta_1<\mu$ and $\eta_2<\mu$, then the tightest lower bound for $\Pro(\eta_1\leq x \leq \eta_2)$ is zero.

In the following section, we show that the proposed upper and lower bounds can improve an estimation of probability of true positives and false positives for a classifier. Based on the results, a closed form parametric classifier is proposed. Similarly, for non-parametric problems, a loss function is derived using an estimation of true and false positives for logits. 

\section{Parametric Classifiers}\label{parametric_class}
Consider a family of pdfs called $\mathcal{F} = \{f(\mathbf{x}, \boldsymbol{\theta}),  \boldsymbol{\theta}\in \Theta\}$, where $\boldsymbol{\theta}$ is the vector of unknown parameters of pdf and $\Theta$ indicates to the set of all possible unknown parameters. We also assume that the following classification problem is given by
$ c_i : \boldsymbol{\theta} \in \Theta_i, i \in \{0,\cdots, N-1\}$. Now, consider a classifier given as $d_n = \mathrm{arg}\max_{n}\limits g_n(\widehat{\theta}(\mathbf{x}))$, where $g_n$ is the corresponding statistic for the $n^{\mathrm{th}}$ class. Class $n$ is selected if $g_n$ is greater than $g_i$ for all $i \neq n$. Moreover, the parameters of $g_n$ are estimated using the pdf of $\mathbf{x}$. In this section, we propose a method to calculate $g_n$, by optimizing the lower bound of probability of true positives. There is a threshold we denote as $\tau_n$, such that the $n^{\mathrm{th}}$ class is selected by the classifier if $g_n(\widehat{\theta}(\mathbf{x})) > \tau_n$. By replacing $\eta_1$ with $\tau_n$ and $\eta_2$ with infinity in (\ref{Theo_1_EQ_15}), we have 
\begin{eqnarray}\label{para_2}\label{para_3}
\nonumber \Pro(d_n = n| c_n) = \Pro_{\boldsymbol{\theta}\in \Theta_n}(g_n(\widehat{\theta}(\mathbf{x}))>\tau_n)
\geq
1 - \frac{\sigma_n^2}{\sigma_n^2 + (\mu_n-\tau_n)^2} =
\frac{s_n}{1 + s_n},
\end{eqnarray}
where $s_n\pardef \frac{(\mu_n-\tau_n)^2}{\sigma_n^2}
\geq 0$, $\mu_n$ is the mean of $g_n(\widehat{\boldsymbol{\theta}}(\mathbf{x}))$, and $\sigma_n^2$ is the variance of 
$g_n(\widehat{\boldsymbol{\theta}}(\mathbf{x}))$ which are functions of $\boldsymbol{\theta}$. 

Next, we maximize the right hand side of (\ref{para_3}) with respect to $g_n$. 
We will apply some assumptions on $\Theta_n$'s and $g_n$'s to extract a closed form for $g_n$. 
We assume that $g_n$ is both differentiable and non-negative. 
Note that the latter condition can be relaxed if we admit an arbitrary bias for $g_n$.

Because the right hand side of (\ref{para_3}) is an increasing function of $s_n$ for $s_n>0$,  maximizing $s_n$ is sufficient to find $g_n$. 
We assume the estimation of the unknown parameter denoted by $\widehat{\boldsymbol{\theta}}(\mathbf{x})$ is unbiased, and therefore $\mu_n = g_n(\boldsymbol{\theta})$. 
Furthermore, from the Cramer Rao bound theorem, it can be shown that $\sigma^2_n \geq \frac{\partial^T g_n(\boldsymbol{\theta})}{\partial\boldsymbol{\theta}}\mathbf{I}_{\boldsymbol{\theta}}^{-1}\frac{\partial g_n(\boldsymbol{\theta})}{\partial\boldsymbol{\theta}}$, where $\mathbf{I}_{\boldsymbol{\theta}}$ indicates the Fisher Information Matrix and $^T$ is the transpose operator. Hence, to maximize $s_n$, the denominator should reach to 
$\frac{\partial^T g_n(\boldsymbol{\theta})}{\partial\boldsymbol{\theta}}\mathbf{I}_{\boldsymbol{\theta}}^{-1}\frac{\partial g_n(\boldsymbol{\theta})}{\partial\boldsymbol{\theta}}$. This means that $g_n$ is given by solving the following optimization problem
\begin{eqnarray}\label{maxim_1}
g_n = \mathrm{arg}\max_{g_n}\limits \frac{(g_n(\boldsymbol{\theta})-\tau_n)^2}{\frac{\partial^T g_n(\boldsymbol{\theta})}{\partial\boldsymbol{\theta}}\mathbf{I}_{\boldsymbol{\theta}}^{-1}\frac{\partial g_n(\boldsymbol{\theta})}{\partial\boldsymbol{\theta}}}
\end{eqnarray}
To solve this problem, we define a new parameter called $\boldsymbol{\vartheta}\pardef \boldsymbol{\vartheta}(\boldsymbol{\theta})$ as a function of $\boldsymbol{\theta}$, such that, 
\begin{eqnarray}\label{helper_0}
\frac{\partial^T \boldsymbol{\vartheta}(\boldsymbol{\theta})}{\boldsymbol{\partial\theta}}\mathbf{I}_{\boldsymbol{\theta}}^{-1}\frac{\partial \boldsymbol{\vartheta}(\boldsymbol{\theta})}{\partial\boldsymbol{\theta}} = \mathds{I},
\end{eqnarray}
where $\mathds{I}$ is the identity matrix. Through utilization of the chain rule and (\ref{helper_0}), the optimization problem in (\ref{maxim_1}) becomes 
\begin{eqnarray}\label{maxim_2}
g_n = \mathrm{arg}\max_{g_n}\limits \frac{(g_n(\boldsymbol{\vartheta})-\tau_n)^2}{\|\frac{\partial g_n(\boldsymbol{\vartheta})}{\partial\boldsymbol{\vartheta}}
\|^2}
\end{eqnarray}
To further reduce (\ref{maxim_2}), we consider contours of $g_n(\boldsymbol{\vartheta})= c$. 
We can solve (\ref{maxim_2}) to extract the shape of contours. By defining $j \pardef (c-\tau_n)^2$, we arrive at 
\begin{eqnarray}\label{maxim_3}
g_n = \mathrm{arg}\min_{g_n}\limits \frac{1}{j}\|\frac{\partial g_n(\boldsymbol{\vartheta})}{\partial\boldsymbol{\vartheta}}
\|^2.
\end{eqnarray}
The objective function of (\ref{maxim_3}) is invariant with respect to $\mathbf{U}\boldsymbol{\vartheta}$, where $\mathbf{U}$ is an arbitrary unitary matrix. 
Therefore, $g_n$ is an invariant function with respect to $\mathbf{U}\boldsymbol{\vartheta}$. Note that any invariant function is a function of the maximal invariant. 
In this case, the maximal invariant of $\mathbf{U}\boldsymbol{\vartheta}$ is $\|\boldsymbol{\vartheta}\|^2$. 
A maximal invariant is a function that satisfies two conditions. 
First, that it is invariant, written $\|\mathbf{U}\boldsymbol{\vartheta}\|^2 = \|\boldsymbol{\vartheta}\|^2$. 
Second, if the maximal invariant for two vectors is equal, there exists a transformation between them. In this case, if $\|\boldsymbol{\vartheta}_0\|^2 = \|\boldsymbol{\vartheta}_1\|^2$, there is a unitary matrix $\mathbf{U}$ such that $\boldsymbol{\vartheta}_0 = \mathbf{U}\boldsymbol{\vartheta}_1$. 
As the optimal function is a function of $\|\boldsymbol{\vartheta}\|^2$, its contours are given by $\|\boldsymbol{\vartheta}\|^2$. 
Therefore, a function that satisfies (\ref{maxim_3}) is $\|\boldsymbol{\vartheta}\|^2$, where $\boldsymbol{\vartheta}$ is given by solving the partial differential problem (\ref{helper_0}). 
Using this, we can apply a boundary condition to get a specific result. 
A simple boundary condition could be that $\|\boldsymbol{\vartheta}\|^2=0$ for at least $\boldsymbol{\theta}\in \partial \Theta_n$, where $\partial \Theta_n$ denotes the boundary of set of $\Theta_n$. The following example shows a method to find $\boldsymbol{\vartheta}$ and its corresponding classifier. There are practical examples that show that the proposed method outperforms the estimation of likelihood function. In the following, we show a toy example to show how to extract $g_n$. 
\subsection{Example}
Consider a binary random variable denoted by $x_i$, which has the value $1$ with a probability $\theta_i$ and $0$ with a probability $1- \theta_i$. Assume that we observe a vector $\mathbf{x}$, composed of such $x_i$. 
Further, assume we have a binary classification problem, such that for the first class $\theta_i< p_i$ and for the other class $\theta_i\geq p_i$. The Fisher information matrix for this pdf is a diagonal matrix where the $i^{\mathrm{th}}$ element of its diagonal is $\frac{1}{\theta_i(1-\theta_i)}$. 
Moreover the maximum likelihood estimation of $\theta_i$, denoted by $\widehat{\theta_i}$, is $1$ if $x_i = 1$ and $p_0$ when $x_i = 0$. From (\ref{helper_0}), we have $\vartheta_i = \sin^{-1}(\theta_i) + k_i$, where $k_i$ is a constant. 
We can set $k_i = \sin^{-1}(p_i)$ to hold the boundary conditions. 
By substituting $\vartheta_i$ with $g_1= \sum_{i}\vartheta_i^2$ and also substituting $\widehat{\theta_i}$ for $\theta_i$, we arrive at $g_1 = \sum_{i}\cos^{-1}(p_i)x_i$. 
This result is a linear classifier, weighted by $\cos^{-1}(p_i)$. 
Along a different line, the maximum likelihood ratio of this problem, found by replacing the estimation of unknown parameters with maximum likelihood estimations, is another linear classifier, namely $g_{\mathrm{ML}} = -\sum_{i}\log(p_i)x_i$. 

We compare the accuracy of the proposed classifier with the estimation of likelihood ratio for this problem. The simulation result shows the proposed classifier outperforms the estimation of likelihood ratio. Consider a vector $\mathbf{x}$ with size 10, and also $p_i=0.001$ for $i = 1,2,3,4$ and $p_i=0.1$ for $i = 5,6,7,8,9,10$. The numerical simulation shows that, the accuracy of the proposed classifier is $0.87$ while the accuracy of the estimation of likelihood ratio is $0.85$.

\section{Non-parametric Classifier}\label{non_par}
When the pdf of data is unknown and a labeled dataset is available, we can apply a training strategy based on the proposed lower bound of probability given by (\ref{Theo_1_EQ_15}). Assume $x_n$ is the output of a classification model, where the model returns class $n$ if $x_n>x_i$ for all $i\neq n$. Such an $x_n$ is the logit for the $n^{\mathrm{th}}$ class in a classification model. 
If we apply (\ref{Theo_1_EQ_15}) for each logit, we find
\begin{eqnarray}\label{prob_logits_nn_0}
\Pro(d = n|c_n) = \Pro(x_n>\eta_n|c_n)\geq
1 - \frac{\sigma_n^2}{\sigma_n^2 + (\mu_n-\eta_n)^2} =\frac{(\mu_n-\eta_n)^2}{\sigma_n^2 + (\mu_n-\eta_n)^2}=\frac{s_n}{1 + s_n},
\end{eqnarray}
Moreover, the other logits should be less than $\eta_n$,
\begin{eqnarray}\label{prob_logits_nn_1}
\Pro(x_i<\eta_n|c_n)\geq \frac{(\eta_n-\mu_{i|n})^2}{(\eta_n-\mu_{i|n})^2 + \sigma^2_{i|n}} = \frac{s_{i|n}}{1 + s_{i|n}},
\end{eqnarray}
where the right hand side of (\ref{prob_logits_nn_1}) is given by (\ref{CDF_lower_0}). 
In (\ref{prob_logits_nn_1}), we define $s_{i|n} \pardef \frac{(\eta_n-\mu_{i|n})^2}{\sigma^2_{i|n}}$ and $s_{n} \pardef \frac{(\eta_n-\mu_n)^2}{\sigma^2_n}$, where $\mu_{i|n}$ and $\sigma^2_{i|n}$ are the conditional mean and variance of $i^{\mathrm{th}}$ logit, respectively, given $c_n$ is the input data.
Similarly, $\mu_{n}$ and $\sigma^2_{n}$ are the conditional mean and variance of $n^{\mathrm{th}}$ logit. Based on the necessary conditions of (\ref{CDF_lower_0}) and (\ref{Theo_1_EQ_15}), we note that $\eta_n>\mu_{i|n}$ and $\eta_n<\mu_{n}$. We designate $s_n$ and $s_{i|n}$ as the intra-class and inter-class Signal to Noise Ratios (SNRs). To improve the accuracy of such a model, the left hand side of (\ref{prob_logits_nn_0}) and (\ref{prob_logits_nn_1}) should be maximized. 
We do so by maximizing the lower bound we have formulated for it. 
As the right hand side of (\ref{prob_logits_nn_0}) and (\ref{prob_logits_nn_1}) are increasing functions of $s_{n}$ and $s_{i|n}$, maximizing them provides the same results as maximizing the lower bounds. Therefore, we propose the following loss function for the $n^{\mathrm{th}}$ and $i^{\mathrm{th}}$ logits
\begin{eqnarray}\label{snr_loss_0}
l_{\mathrm{SNR},n,i} = \frac{1}{s_n} + \frac{1}{s_{i|n}} + \lambda \big(\max(0, \mu_{i|n}-\eta_n-m) + \max(0, \eta_n -\mu_{n}+m)\big),
\end{eqnarray}
where $\lambda$ is the Lagrange multiplier to satisfy the constraints $\eta_n>\mu_{i|n}$ and $\eta_n<\mu_{n}$. To improve the satisfaction of these constraints, a margin $m$ is included. To achieve consistency with loss functions like cross entropy, we choose not to maximize the SNRs, but instead minimize the inverse of them, the Noise to Signal Ratio (NSR). 
Finally, the general loss function is given by taking a summation over all logits:
\begin{eqnarray}\label{snr_loss_1}
l_{\mathrm{SNR}} = \sum_{n}\sum_{i\neq n} l_{\mathrm{SNR},n,i}.
\end{eqnarray}
To implement (\ref{snr_loss_1}), we need to initialize the $\eta_n$'s and update them. We apply two strategies to update them. In the first method, we update them for each batch. Each batch may not contain enough data from all classes, therefore we also propose updating the $\eta_n$'s after each epoch. 
In the following section, we will evaluate these two implementations of the proposed loss function, combined with cross entropy. We denote the first implementation as SNR loss batch-wise and the second one as SNR loss epoch-wise.

Consider the $t^{\mathrm{th}}$ batch, then we set a margin $\Delta^{(t)}$ as $\Delta^{(t)} = m\widehat{\sigma_n}^{(t)}$, where $\widehat{\sigma_n}^{(t)}$ is the sample standard deviation which is calculated from the batch and $m$ is a constant. In our implementation, we set it as $4$. Then, we update $\Delta^{(t)}$ as follows $\eta_n^{(t+1)} = \widehat{\mu_n}^{(t)} - \Delta^{(t)}$, where $\widehat{\mu_n}^{(t)}$ is the sample mean for batch $t$. In each batch, we use the updated $\eta_n$ from the previous batch. For the epoch-wise implementation, $t$ indicates the $t^{\mathrm{th}}$ epoch.

\subsection{Implementation}\label{implementation}
To evaluate the effectiveness of the proposed SNR loss function, we measure its classification accuracy on MNIST, CIFAR-10 and CIFAR-100 when used as the loss for a neural network. For MNIST, LeNet \cite{lecun1998gradient} is chosen as the model architecture. For CIFAR-10 and CIFAR-100, ResNet-22 \cite{he2016deep}. In all experiments, we train the networks using momentum stochastic gradient descent \cite{bower2012book} with $\beta = 0.9$, and a batch size of $1024$. We use the summation of Cross Entropy and SNR loss (CE+SNR) as the loss function, and compare it with using solely the Cross Entropy loss (CE).

The classification results on MNIST, CIFAR-10, and CIFAR-100 are shown in Table \ref{result-table}. 
For all three, the addition of the SNR loss function yields a higher test accuracy. 
The proposed loss function with ResNet-22 achieves 92.04 percent classification accuracy on CIFAR-10, a 1.33 percent improvement over cross entropy loss alone. 
Furthermore, our loss function for ResNet-22 on CIFAR-100 improves the accuracy from 62.10 percent to 70.44 percent, which is a non-trivial gain. 
Additionally, computation of the SNR loss does not slow down the training process, as the extra computation 
occurs only at the end of the batch or epoch and is insignificant compared to the large number of sample-wise operations. 

\begin{figure}
  \centering
  \includegraphics[scale=0.27]{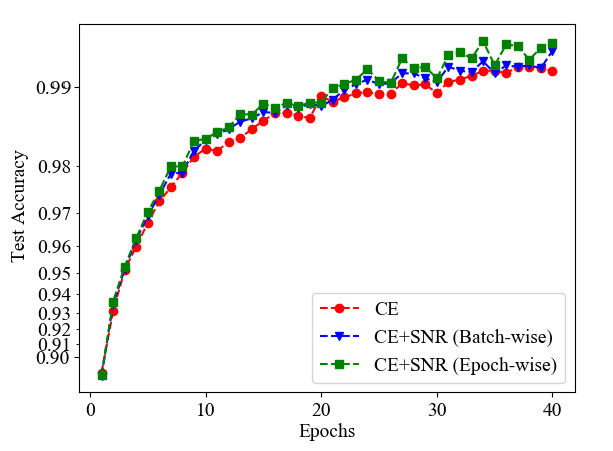}
  \includegraphics[scale=0.28]{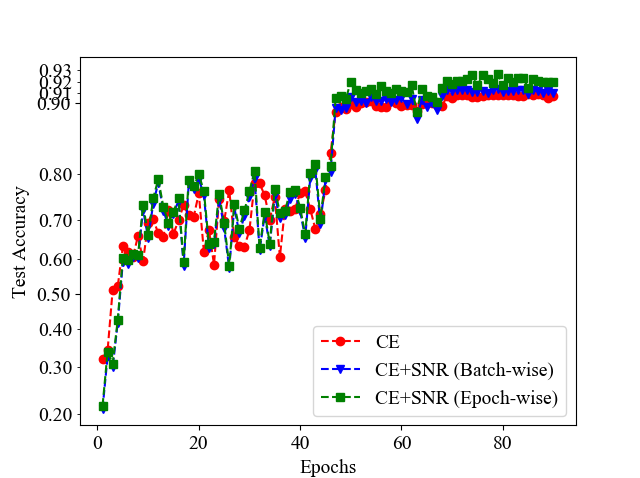}
  \includegraphics[scale=0.28]{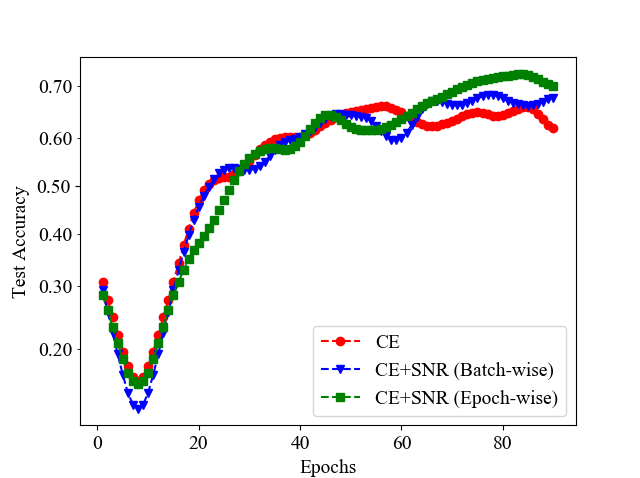}
  \caption{Validation accuracy versus epochs for cross entropy and cross entropy together with two implementations of SNR loss (batch-wise and epoch-wise) on MNIST (left), CIFAR-10 (middle) and CIFAR-100 (right) dataset.}\label{fig_mnist}
\end{figure}

Figure \ref{fig_mnist} shows the validation accuracy of MNIST and CIFAR-10 and CIFAR-100 as a function of the number of epochs. The proposed losses outperform cross entropy on its own, with the difference on MNIST already becoming apparent after only a few epochs. For these three datasets, it can be plainly seen that the SNR loss achieves a similar level of accuracy during the training faster than cross entropy by itself. This means that the proposed loss can improve the speed of training, as well.


\begin{table}
  \caption{Classification results (accuracy \%) on MNIST, CIFAR-10 and CIFAR-100 datasets.}
  \label{result-table}
  \centering
  \begin{tabular}{lllll}  
    \toprule
    Dataset         & Model     & CE    & CE+SNR     & CE+SNR \\
    
                    &           &       &(Batch-wise)  &(Epoch-wise)\\
    \midrule
    MNIST           & LeNet-5       & 99.13   & 99.27       & \textbf{99.32}   \\
    CIFAR-10        & ResNet-22     &90.71    & 91.05       & \textbf{92.04}   \\
    CIFAR-100       & ResNet-22     &66.62   & 67.36       & \textbf{70.44}   \\
    \bottomrule
  \end{tabular}
\end{table}

%

\section{Conclusion}
Within this work, we have identified an overlooked source of information available when training a classifier, and have developed the framework necessary to exploit that information. To that end, we have derived bounds for the probability of true and false positives for parametric classification problems, and provided a closed form for maximizing these bounds. We have also formulated a loss function to optimize these bounds for non-parametric problems. The proposed loss has improved performance across several datasets and algorithms, and suggests broad applicability within the domain of classification with minimal computational cost. 

\bibliographystyle{plainnat}
\bibliography{main}

\end{document}